\pdfoutput=1
\documentclass{article}

\usepackage[
  a4paper,
  left=20mm,
  right=20mm,
  top=25mm,
  bottom=25mm
]{geometry}

\usepackage{graphicx}

\usepackage{graphicx}
\usepackage{authblk}
\usepackage{cite}
\usepackage{amsmath,amssymb,amsfonts}
\usepackage{algorithmic}
\usepackage{graphicx}
\usepackage{textcomp}
\usepackage{placeins}

\title{Personalized and Explainable Blood Pressure Estimation from PPG via Hybrid CNN--Morphological Features}

\author[1]{Myung-Kyu Yi}
\author[1]{Jongshill Lee}
\author[1]{Jeyeon Lee}
\author[1]{In Young Kim}

\affil[1]{%
Department of Biomedical Engineering, Hanyang University,\\
Seoul 04763, South Korea\\
\texttt{\{kainos14, netlee, jeyeonlee, iykim\}@hanyang.ac.kr}
}

\date{}

\begin{document}

\maketitle

\begin{center}
\begin{minipage}{0.9\textwidth}
\footnotesize
\textbf{Funding:}
This work was supported by the Institute of Information \&
Communications Technology Planning \& Evaluation (IITP) grant funded by
the Korean government (MSIT) (No.~RS-2024-00357879, AI-Based Biosignal
Fusion and Generation Technology for Intelligent Personalized Chronic
Disease Management).
\end{minipage}
\end{center}

\begin{abstract}
Continuous cuffless blood pressure (BP) monitoring using photoplethysmography (PPG) offers a promising solution for personalized healthcare. However, existing methods have two major limitations. Handcrafted feature-based approaches rely on precise fiducial point detection and are limited to short-term analysis, while deep learning models, despite their accuracy, often operate as black boxes with limited physiological interpretability. To address these challenges, we propose a physiology-guided hybrid framework for personalized BP estimation that couples a convolutional neural network (CNN) branch—capturing global and local waveform dynamics—with a morphology-prior branch that explicitly encodes person-specific vascular characteristics. By embedding a morphology-based feature set that explicitly encodes individual vascular characteristics, the proposed framework enhances personalization and reduces dependence on large-scale training datasets. Evaluated on a subset of the MIMIC-III database under a subject-specific (personalized) testing protocol, the proposed personalized physiology-guided hybrid approach achieved mean absolute errors (MAEs) of $3.77 \pm 0.50$ mmHg for systolic BP and $2.36 \pm 0.40$ mmHg for diastolic BP, corresponding to relative improvements of 43.7\% and 32.4\% over a subject-specific (personalized) CNN-only baseline. SHAP-based analysis confirmed that the introduced morphology-prior features align with individual vascular characteristics, reinforcing per-subject interpretability. These findings highlight the potential of personalized, physiology-guided hybrid learning with novel morphological descriptors for accurate and explainable BP monitoring in real-world settings.
\end{abstract}

\section{Introduction}

Hypertension is a leading risk factor for cardiovascular morbidity and mortality worldwide. The 2024 WHO Global Hypertension Status Report states that about 1.34 billion adults, or 32.5\% of the global population, are hypertensive~\cite{who2024}. This accounts for over 10.4 million deaths annually and even surpasses the combined mortality from all infectious diseases~\cite{ncdcountdown2024, who2024}. Without substantial intervention, this number could rise to 1.68 billion by 2040~\cite{ncdcountdown2024, zhou2023}. Despite broad clinical and public health efforts, global rates of blood pressure (BP) control remain inadequate. This underscores the urgent need for improved monitoring strategies~\cite{gbd2023, roth2023}. However, the current clinical standard—cuff-based sphygmomanometry—is inherently intermittent. While its simplicity and validated accuracy in static conditions explain its popularity, it captures only brief snapshots, missing the dynamic fluctuations and diurnal patterns essential for timely intervention~\cite{mukkamala2023, obrien2024}. These limitations hinder effective management. BP fluctuates in response to circadian rhythms and daily activities. Short office readings can miss conditions such as white-coat hypertension (27\%), masked hypertension (19\%), and abnormal nocturnal dipping patterns (up to 68\%)~\cite{stergiou2024, kario2023}. These diagnostic blind spots are linked to higher cardiovascular risk and poorer outcomes. To address these challenges, guidelines from the ESC and ESH advocate for a shift toward continuous BP monitoring~\cite{williams2024}. 

In contrast to conventional cuff-based methods, continuous monitoring captures full circadian variability and transient fluctuations, enabling comprehensive 24-hour BP profiling, which enhances risk stratification and facilitates timely therapeutic adjustments. Advances in wearable sensing and signal processing have accelerated the development of non-invasive, cuffless solutions for real-world settings~\cite{olsen2024, mukkamala2023}, with photoplethysmography (PPG) being a primary enabling technology. Owing to its low cost, convenience, and seamless integration into consumer devices such as smartwatches and smartphones, PPG has emerged as a promising modality for continuous, real-world BP monitoring. Early cuffless approaches relied on pulse transit time (PTT) or pulse wave velocity (PWV) derived from  electrocardiogram (ECG) and PPG signals, but often needed multiple sensors and frequent calibration, limiting practical adoption. Recent studies have increasingly explored PPG-only solutions that leverage morphological analysis and machine learning to continuously estimate BP. Advanced approaches utilize transfer learning and graph-based feature mapping to extract PPG dynamics efficiently, enabling accurate BP estimation even from short-duration segments~\cite{haddad2022, wang2022, wu2024}.

Despite notable advances in non-invasive BP monitoring technologies, existing PPG-based methods for personalized BP estimation still face critical limitations. Many approaches depend on detecting specific fiducial points, such as the systolic peak or dicrotic notch, for feature extraction, making them highly susceptible to errors when signal quality is degraded by motion artifacts, sensor displacement, or other noise sources. To avoid this dependency, numerous non-fiducial studies have instead relied on statistical properties of the signal or general indices derived from signal processing. While these descriptors can provide some predictive value, they often lack clear physiological interpretability and exhibit only weak correlations with hemodynamic parameters, thereby limiting their physiological validity. Moreover, most feature-based methods extract such indices from short segments covering only a few cardiac cycles, which constrains their ability to capture long-term BP dynamics that are essential for personalized modeling. End-to-end deep learning approaches overcome some of these issues by directly processing continuous waveform data and learning temporal variability, yet the learned features are typically black-box representations that offer little physiological transparency, undermining clinical trust. In addition, these models require large-scale labeled datasets and substantial computational resources to achieve reliable performance, further restricting their applicability in real-world personalized healthcare scenarios.

To overcome these challenges, we propose a hybrid personalized model explicitly designed for per-subject modeling, jointly learning from raw PPG and physiology-aware features. Unlike conventional hybrid approaches that rely on geometric descriptors (e.g., width, amplitude), our method shifts the focus toward hemodynamic features that quantify the forces driving blood flow, such as flow energy and vascular resistance. By utilizing integral-based (area) and rate-based computations rather than point-wise fiducial detection, the proposed feature set inherently suppresses high-frequency noise, reduces sensitivity to small errors in peak or notch localization, and captures holistic waveform dynamics over the entire cardiac phase. These hemodynamic priors are integrated into a lightweight CNN architecture, ensuring that the model captures both local temporal variability and global physiological constraints without the computational burden of heavy deep learning models. Furthermore, normalized feature formulations suppress inter-individual variability, enabling per-subject adaptation without additional calibration. By unifying these components, the framework mitigates several limitations of existing hybrid approaches and provides a compact and interpretable basis for non-invasive continuous BP estimation.

The main contributions of this work are as follows:

\begin{itemize}
    \item We introduce a novel set of ten morphological features that transition the analytical focus from geometric description to the interpretation of underlying vascular physical properties. Unlike noise-sensitive fiducial descriptors, our integral- and rate-based formulation ensures robustness while directly capturing hemodynamic mechanics (e.g., stiffness, compliance) to enhance personalization and minimize calibration needs, as validated by SHAP analysis.

    \item We integrate the proposed features as subject-specific physiological priors into a personalized hybrid convolutional neural network (CNN)–based regression model, combining the strengths of raw waveform modeling with interpretable morphological representations. Under a personalized evaluation protocol, this architecture achieves substantial gains—reducing MAE for systolic BP (SBP) from 6.70 to 3.77 mmHg and for diastolic BP (DBP) from 3.49 to 2.36 mmHg—corresponding to 43.7\% and 32.4\% improvements over the CNN-only baseline. SHAP analysis highlights Post-Notch Recovery Slope (PNRS), Sum of Upstroke Amplitude (SUA), and Tail Entropy (TE) as dominant contributors, strengthening physiological explainability.
\end{itemize}

The remainder of this paper is organized as follows. Section II reviews prior research on cuffless and personalized BP estimation using PPG signals. Section III describes the proposed methodology. Section IV presents the experimental setup, results, and SHAP-based interpretability analysis. Section V discusses the physiological interpretation of feature importance and the implications for clinical applicability. Section VI concludes with key findings and future directions.

\section{Related Work}

Recent advances in cuffless blood pressure (BP) estimation from photoplethysmography (PPG) have progressed from morphology-based handcrafted features, to end-to-end deep learning on raw waveforms, and, most recently, to hybrid approaches that fuse interpretable morphological priors with data-driven representations to improve accuracy and generalizability under dynamic conditions. In the early stages of this research, securing physiological interpretability was paramount. Investigators focused on manually extracting morphological features from PPG waveforms—such as systolic upstroke time, diastolic area ratios, and especially pulse transit time (PTT). Because these handcrafted features are directly linked to physiological factors like vascular elasticity and peripheral resistance, they formed the basis of traditional models \cite{liu2023morph, wang2023dual} and provided an important interpretive foundation.

As deep learning matured and large-scale datasets became available, the field’s center of gravity shifted toward data-driven learning. Early deep-learning studies leveraged abundant data to train regression models such as CNNs and LSTMs that ingest the entire raw PPG waveform and allow the model to automatically capture its complex, nonlinear relationship with BP \cite{Cui2025JBHI, wang2022priorlstm}. This line of work aimed to achieve high predictive performance without manual feature engineering, and has recently expanded to multi-dimensional representations—for example, converting 1-D PPG into 2-D images for CNNs \cite{kim2023dim2d}, or transforming the waveform and its derivatives into time–frequency images for analysis \cite{koparir2024cnn}.

Hybrid and explainable-AI (XAI) approaches have emerged to combine the strengths—and overcome the limitations—of the two prior paradigms. These contemporary studies emphasize not only performance but also physiological readability and data efficiency. For instance, methods like MTFF \cite{MTFF} internalize feature engineering by enabling the network itself to learn and fuse morphological, spectral, and temporal cues. In addition, physics-informed models such as PINNs and PITNs \cite{PINN, PITN} incorporate physiological constraints directly into the loss function, enhancing model stability in data-scarce settings. Furthermore, hybrid deep frameworks like HGCTNet \cite{HGCTNet} integrate interpretable handcrafted features (e.g., PTT) into deep models to complement learned representations. Similarly, Botrugno et al. \cite{botrugno2025optimized} introduced a hybrid CNN-SVR architecture that refines deep learning predictions derived from multiwavelength PPG by employing a secondary support vector regression stage to explicitly incorporate physiological covariates. Collectively, these efforts reflect a concerted push to combine morphological knowledge with deep feature learning to achieve both strong performance and physiological interpretability. However, hybrid and explainable approaches often remain at a global level, offering little subject- or segment-specific, quantitative attribution and rarely tying signed feature contributions to established vascular mechanisms—limitations that impede clinical trust.

More recently, research attention has shifted toward personalization strategies to address inter-subject variability—a key barrier in deploying BP models in real-world scenarios. Chakraborty et al.\cite{chakraborty2023personalized} introduced a lightweight, calibration-based framework that leverages peak-to-peak amplitude and foot-to-foot delay, without requiring secondary signals such as ECG, targeting IoT-enabled personalized healthcare systems. Leitner et al.\cite{leitner2022} demonstrated that transfer learning on hybrid CNN–RNN architectures significantly reduces the need for subject-specific data, achieving SBP/DBP MAEs of 3.52/2.20 mmHg using only 50 samples. Building upon this, Fan et al. \cite{fan2023fewbp} proposed FewShotBP, a few-shot transfer learning method that combines multimodal spectro-temporal feature learning with a personalization adapter (PA), enabling effective adaptation with as few as 5–10 labeled samples while minimizing computational overhead—critical for wearable deployments. Zhang et al.\cite{zhang2025personalized} addressed long-term monitoring challenges in wearable scenarios by introducing SCI-GTCN, a lightweight gated temporal convolutional network integrated with adaptive calibration and signal quality grading, demonstrating robustness under circadian fluctuations and rapid BP transitions.

While recent advancements have been made in cuffless BP estimation, challenges remain for real-world application. Conventional feature-based approaches are highly sensitive to motion artifacts and inter-subject variability, often leading to unstable predictions. Conversely, end-to-end deep learning models, while capable of learning complex temporal patterns, typically require impractical amounts of labeled data and frequent recalibration. To address these issues, hybrid approaches combining handcrafted features with deep learning have been explored; however, they predominantly rely on conventional morphological features, as summarized in the middle column of Table~\ref{tab:comparison}. These descriptors focus on the geometric characteristics of the PPG waveform, such as amplitude and width, and depend heavily on the precise detection of fiducial points, making them vulnerable to noise and limiting their direct connection to blood pressure regulation. In contrast, our proposed framework introduces hemodynamic-motivated features (Table~\ref{tab:comparison}, right column) that complement geometric descriptors with simple hemodynamic considerations. These features are designed to reflect vascular properties such as stiffness, compliance, and effective vascular load through energy- and decay-related indices, providing a more physiology-guided association with the cardiovascular system. By relying on area- and rate-based computations over systolic and diastolic phases, the proposed features reduce dependence on specific signal points and improve robustness to fiducial errors and high-frequency artifacts. Integrating these hemodynamic priors into a lightweight CNN architecture helps the model capture both local waveform dynamics and higher-level physiological trends, while mitigating some of the black-box concerns associated with purely end-to-end deep models.

\begin{table*}[!htbp]
\centering
\scriptsize
\caption{Comparison between conventional morphological features and the proposed hemodynamic features. The proposed features are designed to complement geometric descriptors by incorporating simple hemodynamic considerations.}
\label{tab:comparison}
\renewcommand{\arraystretch}{1.5}
\begin{tabular}{l | p{0.35\linewidth} | p{0.35\linewidth}}
\hline
\textbf{Comparison Criteria} 
& \textbf{Conventional Morphological Features} 
& \textbf{Proposed Hemodynamic Features} \\ 
\hline
\textbf{Domain} 
& Geometric Signal Description 
& Vascular and Hemodynamic Properties \\ 
\hline
\textbf{Abstraction Level} 
& \textbf{Descriptive:} Quantifies the visible shape of the PPG waveform (width, amplitude, slope). 
& \textbf{Physiology-oriented:} Encodes vascular properties such as stiffness, compliance, and damping through energy- and decay-related indices. \\ 
\hline
\textbf{Mathematical Basis} 
& \textbf{Point-wise:} Heavily relies on fiducial points (e.g., peak, valley), making it sensitive to noise and localization errors. 
& \textbf{Area/Rate-based:} Uses region-level integrals and decay rates over systolic/diastolic phases, improving robustness to fiducial errors and high-frequency artifacts. \\ 
\hline
\textbf{Link to BP} 
& \textbf{Empirical Association:} Uses waveform shape changes that correlate with BP but are only weakly tied to specific regulatory mechanisms. 
& \textbf{Physiology-guided Association:} Features are constructed to reflect determinants of BP—such as arterial stiffness, vascular compliance, and peripheral resistance—rather than geometry alone. \\ 
\hline
\end{tabular}
\end{table*}

\FloatBarrier

\section{Proposed Methodology}

This section introduces a personalized BP estimation pipeline using PPG signals. The framework comprises three primary components:  
(1) Signal preprocessing to remove artifacts and segment the waveform into analyzable units.  
(2) Computation of physiologically interpretable features that capture morphological and hemodynamic properties of the PPG waveform, and  
(3) a hybrid CNN–Prior regression model integrates raw PPG waveform representations with handcrafted PPG feature descriptors for robust BP estimation.

\subsection{Signal Preprocessing}

Raw PPG signals are prone to baseline drift, motion artifacts, and high-frequency noise. To mitigate these artifacts, a multi-stage preprocessing pipeline was employed. First, a zero-phase Butterworth bandpass filter (0.5–8 Hz) was applied to remove baseline wander and suppress high-frequency noise while preserving cardiac harmonics. Residual low-frequency trends were removed via polynomial detrending. The processed signal was segmented into 10-second non-overlapping windows to capture multiple cardiac cycles. A Signal Quality Index (SQI), computed using skewness and kurtosis, was used to exclude segments with severe artifacts. Valid segments were normalized to zero mean and unit variance to ensure amplitude invariance.

Morphological feature extraction was performed on a cycle-wise basis. Physiological landmarks such as the systolic peak and dicrotic notch were detected using an adaptive peak detection algorithm constrained by heart rate ranges (40–180 bpm). Derivatives were computed using centered finite differences to characterize slopes and curvatures. Based on these landmarks, ten handcrafted features were extracted to capture essential waveform properties including upstroke dynamics, energy distribution, and post-notch recovery patterns.

\subsection{Morphological Feature Design}

In this section, we present a refined set of morphological features for BP estimation from PPG signals. Traditional handcrafted descriptors can broadly be divided into two categories. The first category consists of fiducial-point-based features, which rely on the detection of specific waveform landmarks such as peak amplitude, pulse width, or inter-landmark intervals. The second category includes non-fiducial features, which are typically derived from the entire waveform and are often expressed as statistical moments or frequency-domain metrics. While useful, these features remain limited: fiducial-based ones are highly noise-prone and sensitive to landmark detection errors, whereas non-fiducial ones often lack clear physiological interpretability. Moreover, both categories are typically amplitude-dependent and show poor generalizability across subjects and devices.

To overcome these shortcomings, we introduce ten new morphological descriptors designed under four guiding principles:
\begin{enumerate}
\item \textbf{Hemodynamic Relevance:} Features explicitly reflect physiological mechanisms influencing BP, such as arterial stiffness, cardiac contractility, and vascular resistance.
\item \textbf{Global Energy Representation:} Waveform descriptors capture energy distribution across systolic and diastolic phases.
\item \textbf{Differential Shape Encoding:} Sharpness, curvature, and complexity are characterized via differential operators.
\item \textbf{Normalization and Robustness:} Amplitude-independent formulations ensure consistency under varying subjects and signal quality.
\end{enumerate}

This principled design directly addresses the weaknesses of both fiducial and non-fiducial handcrafted features, yielding descriptors that are physiologically interpretable, noise-resilient, and broadly generalizable. We refine the first- and second-order derivatives of the PPG to yield more stable and interpretable shape cues. First derivative descriptors use slope statistics directly. The first-derivative descriptors, defined on $p'(t)$, quantify the rate of change in the PPG signal, reflecting key dynamic events during the cardiac cycle. These include the Sum of Upstroke Amplitude (SUA), calculated as the sum of $|p'(t)|$ over the systolic upstroke~\cite{liu2023morph, wang2022, wang2023dual}; the Post-Notch Recovery Slope (PNRS), which is the mean of $p'(t)$ over the diastolic segment~\cite{liu2023morph ,wang2022, haddad2022}; and the Early Acceleration Ratio (EAR), defined as the ratio between the early-systolic and whole-systolic means of $|p'(t)|$~\cite{liu2023morph,wang2022,MTFF}. Furthermore, the Slope Recovery Ratio (SRR) utilizes finite-difference slopes over diastole, which is equivalent to segment-wise averages of $p'(t)$ \cite{MTFF}, \cite{liu2024symmetrical}. The second-derivative descriptors, defined on $p''(t)$, formalize the waveform's curvature, which is highly sensitive to arterial stiffness and reflected waves. Key second-derivative features are the Second-Derivative Energy (SDE), calculated as $\sum p''(t)^2$ (representing the curvature energy)~\cite{liu2023morph, wang2022}; the Peak Sharpness Index (PSI), which is the local mean of $|p''(t)|$ near the systolic peak~\cite{liu2023morph, wang2022, haddad2022}; and the Waveform Complexity Count (WCC), which counts the zero-crossings of $p''(t)$—the inflection points of $p(t)$ \cite{liu2023morph, haddad2022}. Together, these measures provide a comprehensive characterization of the PPG waveform's slope and curvature patterns, which are essential inputs for cardiovascular health modeling. 

\subsection{Morphological Feature Computation and Interpretation}

\begin{figure*}[!htbp]
    \centering
    \includegraphics[width=0.5\linewidth]{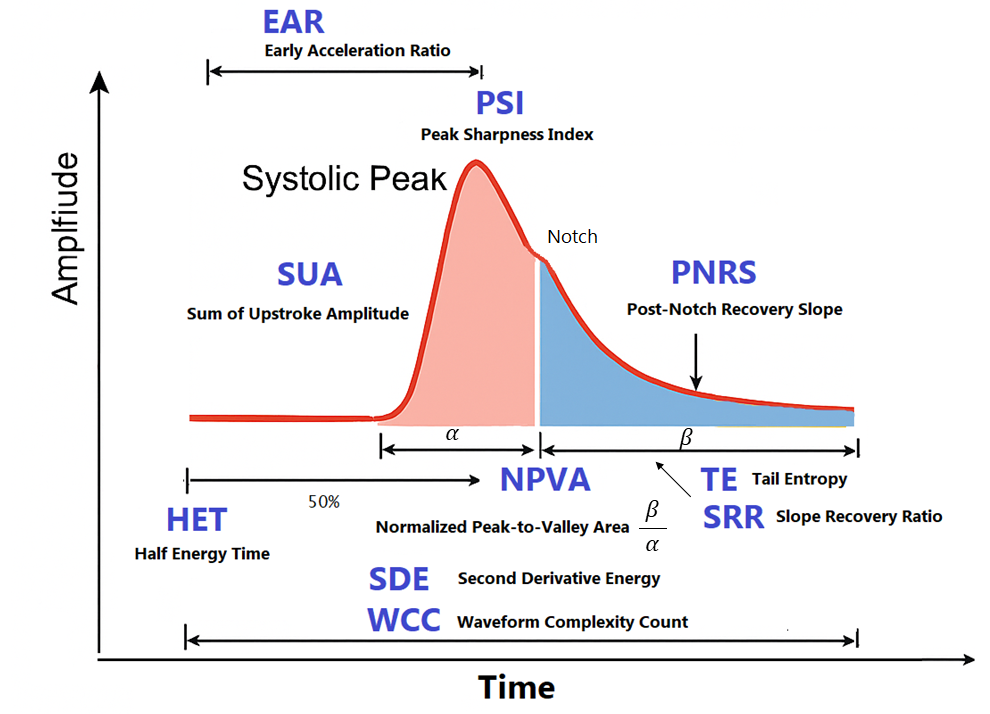}
    \caption{
        Schematic illustration of the proposed morphological features extracted from a single PPG cycle}
    \label{fig:ppg_features}
\end{figure*}

As shown in Fig.~\ref{fig:ppg_features} and Table~\ref{tab:custom_features}, these features capture key temporal and structural properties of the cardiac cycle, providing clinically relevant indicators for vascular compliance, arterial stiffness, and hemodynamic status.

\begin{table*}[!htbp]
\centering
\footnotesize
\caption{Proposed Features with Mathematical and Physiological Interpretations and Design Principles}
\label{tab:custom_features}
\begin{tabular}{l|p{5.5cm}|p{5.5cm}|p{3.5cm}}
\hline
\textbf{Feature} & \textbf{Mathematical Meaning} & \textbf{Physiological Interpretation} & \textbf{Design Principle} \\
\hline
SUA & Accumulated first derivative from onset to systolic peak & Proxy for stroke volume and arterial compliance & Hemodynamic Relevance \\
\hline
PNRS  & Mean first derivative after dicrotic notch & Reflective wave slope, indicator of arterial stiffness & Hemodynamic Relevance \\
\hline
NPVA & Area difference before and after notch, normalized by total area & Balance of systolic and diastolic energy & Global Energy Representation \\
\hline
SDE & Sum of squared second-order derivatives & Signal roughness; reflects vascular tension or resistance & Differential Shape Encoding \\
\hline
TE & Shannon entropy of post-notch segment & Morphological irregularity in diastolic tail & Global Energy Representation \\
\hline
HET & Time to reach 50\% of total energy & Symmetry of energy distribution; waveform temporal centroid & Global Energy Representation \\
\hline
PSI & Local curvature at systolic peak via second derivative & Peak separation quality; vascular tone indicator & Differential Shape Encoding \\
\hline
SRR & Normalized slope between notch and waveform end & Post-notch recovery speed; vascular damping behavior & Normalization and Robustness \\
\hline
WCC & Count of second derivative sign changes & Morphological complexity; inflection richness & Differential Shape Encoding \\
\hline
EAR & Ratio of early to full upstroke slopes & Initial ventricular ejection speed; cardiac contractility marker & Hemodynamic Relevance \\
\hline
\end{tabular}
\end{table*}

\FloatBarrier

\subsubsection*{1) Sum of Upstroke Amplitude (SUA)}
\textit{Computation:}
\begin{align}
\text{SUA} = \sum_{i=0}^{\text{peak}} \left| \frac{d}{dt} \text{PPG}(i) \right|
\end{align}
\textit{Interpretation:} This feature represents the cumulative gradient from the waveform onset to the systolic peak. It serves as a proxy for the overall upstroke energy and is associated with cardiac output and arterial compliance. Higher SUA values may indicate increased contractile force or vascular stiffness. Since both stroke volume and vascular elasticity directly influence SBP, SUA provides an integrated marker for systolic pressure regulation.

\vspace{0.5em}
\subsubsection*{2) Post-Notch Recovery Slope (PNRS)}
\textit{Computation:}
\begin{align}
\text{PNRS} = \frac{1}{N - n} \sum_{i=n}^{N} \frac{d}{dt} \text{PPG}(i)
\end{align}
\textit{Interpretation:} PNRS is the mean first derivative after the dicrotic notch, reflecting the slope of the reflective wave. It is sensitive to peripheral vascular resistance and wave reflection, making it an indirect indicator of arterial stiffness. Greater arterial stiffness elevates wave reflection amplitude, contributing to SBP augmentation.

\vspace{0.5em}
\subsubsection*{3) Normalized Peak-to-Valley Area (NPVA)}
\textit{Computation:}
\begin{align}
\text{NPVA} = \frac{A_{\beta} - A_{\alpha}}{A_{\alpha} + A_{\beta}}
\end{align}
where:
\begin{itemize}
    \item $A_{\alpha}$: Area under the curve in the systolic upstroke region ($\alpha$ phase, before notch)
    \item $A_{\beta}$: Area under the curve in the diastolic decay region ($\beta$ phase, after notch)
\end{itemize}

\textit{Interpretation:} NPVA quantifies the relative area distribution between the diastolic ($\beta$) and systolic ($\alpha$) phases, normalized by the total waveform area. A higher NPVA indicates a greater proportion of diastolic energy relative to systolic energy, potentially reflecting enhanced arterial compliance or altered peripheral resistance. Shifts in this ratio affect diastolic pressure maintenance and pulse pressure.

\vspace{0.5em}
\subsubsection*{4) Second Derivative Energy (SDE)}
\textit{Computation:}
\begin{align}
\text{SDE} = \sum_{i} \left( \frac{d^2}{dt^2} \text{PPG}(i) \right)^2
\end{align}
\textit{Interpretation:} SDE, the sum of squared second derivatives, captures abrupt morphological transitions. It is useful for characterizing vascular tension and detecting waveform irregularities. High vascular tension shortens systolic ejection and elevates SBP, making SDE a surrogate for arterial load.

\vspace{0.5em}
\subsubsection*{5) Tail Entropy (TE)}
\textit{Computation:}
\begin{align}
\text{TE} = H(\text{PPG}_{i>\text{notch}})
\end{align}
\textit{Interpretation:} TE measures the Shannon entropy of the post-notch region, representing waveform irregularity in the diastolic tail. Higher TE values may indicate altered hemodynamics or signal disturbances. Irregular diastolic decay patterns can reflect impaired vascular relaxation, influencing DBP stability.

\vspace{0.5em}
\subsubsection*{6) Half Energy Time (HET)}
\textit{Computation:}
\begin{align}
\text{HET} = t : \sum_{i=0}^{t} \text{PPG}(i)^2 \geq 0.5 \cdot E_{\text{total}}
\end{align}
\textit{Interpretation:} HET represents the time point at which cumulative energy reaches 50\% of the total waveform energy. It serves as an indicator of temporal symmetry, evaluating the balance of energy distribution between the early (systolic) and late (diastolic) phases of the PPG waveform, and can highlight pathological deviations in timing. An altered HET can shift the systolic–diastolic pressure balance, impacting both SBP and DBP.

\vspace{0.5em}
\subsubsection*{7) Peak Sharpness Index (PSI)}
\textit{Computation:}
\begin{align}
\text{PSI} = \frac{1}{3} \sum_{i=\text{peak}-1}^{\text{peak}+1} \frac{d^2}{dt^2} \text{PPG}(i)
\end{align}
\textit{Interpretation:} PSI estimates the local curvature at the systolic peak. Sharper peaks often suggest preserved vascular tone, whereas flatter peaks may indicate damping or arterial aging. Vascular tone directly modulates SBP amplitude by influencing forward wave formation.

\vspace{0.5em}
\subsubsection*{8) Slope Recovery Ratio (SRR)}
\textit{Computation:}
\begin{align}
\text{SRR} = \frac{\text{PPG}[-1] - \text{PPG}[\text{notch}]}{(N - \text{notch}) / f_s}
\end{align}
\textit{Interpretation:} SRR captures the amplitude recovery rate after the notch. It reflects vascular damping and post-systolic waveform behavior. Reduced damping elevates late systolic load, contributing to higher SBP.

\vspace{0.5em}
\subsubsection*{9) Waveform Complexity Count (WCC)}
\textit{Computation:}
\begin{align}
\text{WCC} = \sum \left( \text{sign change of } \frac{d^2}{dt^2} \text{PPG} \right)
\end{align}
\textit{Interpretation:} WCC is the number of sign changes in the second derivative, representing waveform complexity. It helps detect morphological variability across cardiac cycles. Greater complexity may reflect unstable vascular regulation, potentially linked to BP variability.

\vspace{0.5em}
\subsubsection*{10) Early Acceleration Ratio (EAR)}
\textit{Computation:}
\begin{align}
\text{EAR} = \frac{\text{mean} \left( \frac{d}{dt} \text{PPG}[0:\text{peak}/4] \right)}{\text{mean} \left( \frac{d}{dt} \text{PPG}[0:\text{peak}] \right)}
\end{align}
\textit{Interpretation:} EAR measures the ratio of early to total upstroke gradients, reflecting the rapidity of ventricular ejection and early systolic dynamics. Rapid early ejection increases the initial pressure rise rate, often elevating SBP in high-contractility states.

\subsection{Personalized Hybrid CNN--Prior Regression Model}

\begin{figure*}[!htbp]
    \centering
    \includegraphics[width=0.9\linewidth]{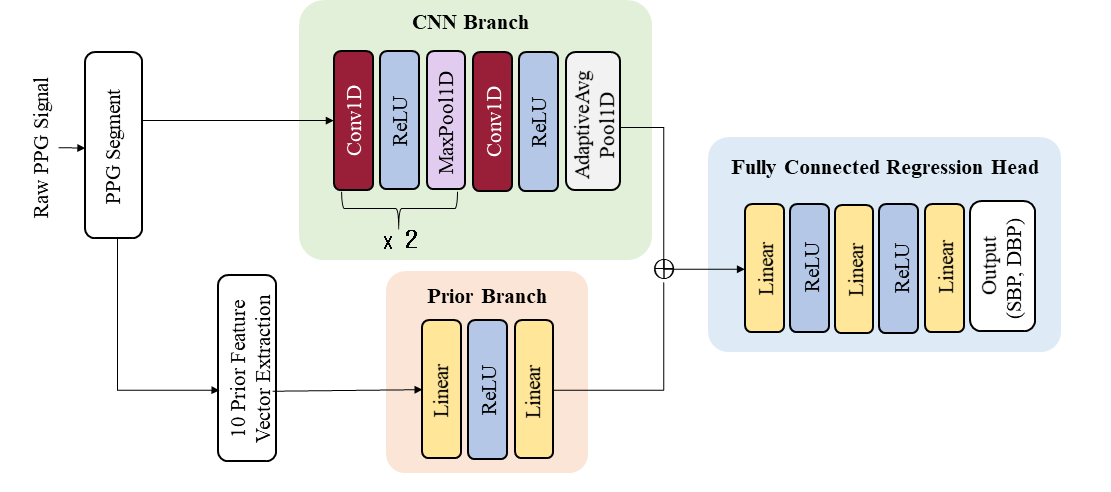}
    \caption{Schematic of the proposed hybrid CNN–Prior model. The architecture comprises two parallel branches: (1) a CNN module that processes raw PPG waveforms to extract local morphological patterns, and (2) a prior branch that encodes handcrafted physiological features. These representations are fused and passed through a fully connected regression head for personalized SBP/DBP prediction.}
    \label{fig:cnn_prior_architecture}
\end{figure*}

To jointly exploit the rich temporal structure of raw PPG signals and the physiological relevance of interpretable morphological descriptors, we propose a hybrid regression model. The model architecture is designed to mitigate the black-box limitations of CNNs by injecting clinically grounded prior knowledge, thereby improving both prediction robustness and interpretability across subjects. This design improves BP estimation accuracy by integrating end-to-end learning with clinically relevant PPG features. As shown in Fig.~\ref{fig:cnn_prior_architecture} the model consists of two branches: a CNN branch for PPG feature extraction and a prior branch for handcrafted feature encoding, followed by a fully connected regression head.

\FloatBarrier

\subsubsection{CNN Branch}
The CNN branch receives the raw waveform input $\mathbf{X} \in \mathbb{R}^{T}$ and processes it through three sequential layers of one-dimensional convolution, non-linear activation $\sigma(\cdot)$, and pooling. Specifically:
\begin{align}
\mathbf{H}_1 &= \text{MaxPool1D}\big(\sigma(\text{Conv1D}_{1 \rightarrow 32}(\mathbf{X}))\big), \\
\mathbf{H}_2 &= \text{MaxPool1D}\big(\sigma(\text{Conv1D}_{32 \rightarrow 64}(\mathbf{H}_1))\big), \\
\mathbf{H}_3 &= \text{AdaptiveAvgPool1D}\big(\sigma(\text{Conv1D}_{64 \rightarrow 128}(\mathbf{H}_2))\big), \\
\mathbf{z}_{\text{cnn}} &= \text{Flatten}(\mathbf{H}_3) \in \mathbb{R}^{128}.
\end{align}
Here, $\mathbf{z}_{\text{cnn}}$ represents the learned deep feature vector capturing fine-grained temporal and morphological variations in the waveform.

\subsubsection{Prior Branch}
The handcrafted morphological prior vector $\mathbf{p} \in \mathbb{R}^{10}$ is passed through two fully connected layers with non-linear activation:
\begin{align}
\mathbf{h}_p &= \sigma(\phi_1(\mathbf{p})), \quad \phi_1: \mathbb{R}^{10} \to \mathbb{R}^{64}, \\
\mathbf{z}_{\text{prior}} &= \phi_2(\mathbf{h}_p), \quad \phi_2: \mathbb{R}^{64} \to \mathbb{R}^{128}.
\end{align}
The resulting vector $\mathbf{z}_{\text{prior}}$ encodes domain-informed descriptors that reflect vascular compliance, arterial stiffness, and contractile dynamics.

\subsubsection{Fusion and Regression Head}
The feature vectors from both branches are concatenated to form
\begin{align}
\mathbf{z} &= [\mathbf{z}_{\text{cnn}}; \mathbf{z}_{\text{prior}}] \in \mathbb{R}^{256}.
\end{align}
This combined representation is fed into a three-layer fully connected regression head:
\begin{align}
\mathbf{h}_1 &= \sigma(\psi_1(\mathbf{z})), \\
\mathbf{h}_2 &= \sigma(\psi_2(\mathbf{h}_1)), \\
\hat{\mathbf{y}} &= \psi_3(\mathbf{h}_2), \quad \hat{\mathbf{y}} = [\hat{y}_{\text{SBP}}, \hat{y}_{\text{DBP}}] \in \mathbb{R}^2,
\end{align}
where $\hat{y}_{\text{SBP}}$ and $\hat{y}_{\text{DBP}}$ denote the predicted SBP and DBP values, respectively.

\subsection{Physiological Mapping of Proposed Morphological Features}

Building on the design principles described earlier, the proposed morphological features are computed from region-level waveform characteristics, minimizing dependency on precise landmark detection and improving robustness under noisy or motion-corrupted conditions. Each feature is explicitly linked to a physiological mechanism involved in BP regulation—stroke volume and arterial compliance (SUA), arterial stiffness and wave reflection (PNRS, PSI), vascular compliance and energy distribution (NPVA, HET), vascular damping (SRR), peripheral resistance and waveform irregularity (TE, WCC), and cardiac contractility (EAR). By capturing these hemodynamic determinants, the feature set provides interpretable inputs that complement the CNN-extracted representations, enabling the hybrid model to achieve both predictive accuracy and clinical explainability.
To further clarify the practical distinction of our method, Table~\ref{tab:feature_comparison} compares the proposed hybrid CNN–prior framework with two commonly used alternatives—fiducial-based handcrafted features and end-to-end deep models—highlighting differences in input representation, interpretability, robustness, data requirements, and typical performance.

\begin{table*}[!htbp]
\centering
\scriptsize
\caption{Comparison of the proposed approach with two common alternatives for PPG-based BP estimation.}
\label{tab:feature_comparison}
\renewcommand{\arraystretch}{1.2}
\begin{tabular}{p{4cm} p{3.5cm} p{3.5cm} p{3.5cm}}
\hline
\textbf{Aspect} & 
\textbf{Fiducial-Based Handcrafted Features} & 
\textbf{End-to-End Deep Models} & 
\textbf{Proposed Hybrid CNN + Physiological Priors} \\
\hline
\textbf{Input Representation} & 
Discrete waveform landmarks  & 
Raw waveform samples & 
Raw PPG + priors \\
\textbf{Physiological Interpretability} & 
Moderate & 
Low  & 
High \\

\textbf{Robustness to Noise / Motion} & 
Low & 
Moderate & 
High \\

\textbf{Data Requirements} & 
Low & 
High & 
Moderate \\

\textbf{Computational Complexity} & 
Low & 
High & 
Moderate \\

\hline
\end{tabular}
\end{table*}

\FloatBarrier

\section{Experimental Results}

\subsection{Dataset and Evaluation Metrics}

A subset of the MIMIC-III waveform database, distributed through PulseDB~\cite{Wang2023PulseDB}, was employed in this study. PulseDB is an open-access benchmark for cuffless BP estimation and includes synchronized PPG, ECG, and arterial BP (ABP) signals sampled at 125~Hz. Each 10-second segment was annotated with the mean SBP and DBP calculated across all cardiac cycles within the segment. To ensure reliable signal quality, segments were evaluated using the skewness-based signal quality index (sSQI). Any 5-second sliding window with sSQI $<$ 0 was removed. Additionally, all PPG signals were bandpass filtered between 0.5–8~Hz using a 4th-order Butterworth bandpass filter to remove baseline drift and high-frequency artifacts. After filtering, each 10-second PPG segment was standardised to zero mean and unit variance on a per-segment basis, and the handcrafted/morphological features were z-score normalised using the statistics of the training split to avoid scale mismatch during feature fusion. Our experiments utilized data from 100 non-hypertensive individuals, including both normotensive and elevated (prehypertensive) subjects. This selection criteria was intentional to align with the proposed system's target application: daily health monitoring and preventative tracking for the general population, rather than clinical management of diagnosed stage 1–2 hypertension or hypertensive crises. Since chronic hypertension induces distinct vascular remodeling (e.g., arterial stiffening) that fundamentally alters PPG morphology compared to non-hypertensive baselines, mixing these distinct populations could confound the validation of the proposed morphological features. Therefore, this study establishes a baseline performance for wellness-focused monitoring, isolating physiological variations within the non-hypertensive (normotensive-to-elevated) population~\cite{leitner2022, zhang2025personalized, leogrande2025dataset}.

Each subject’s data were segmented into non-overlapping 10-s windows (1,250 samples), and each window contained multiple cardiac cycles. Because this study used long-term continuous recordings collected in a daily-life environment from healthy adults and prehypertensive subjects, all segments correspond to natural resting-state conditions, recorded on a single measurement day.  To rigorously prevent data leakage arising from the temporal autocorrelation of physiological signals, we adopted a chronological blocked 5-fold cross-validation protocol with safety buffers, rather than random shuffling. Specifically, each subject’s entire recording was divided into five consecutive, non-overlapping blocks of equal length in strict chronological order. A standard blocked 5-fold procedure was then applied: in each of the five folds, one block served as the test set while the remaining four blocks were used for training, such that every block was used exactly once as the test set across the five folds. Crucially, to eliminate spectral leakage from low-frequency physiological trends (e.g., Mayer waves, thermoregulation, or slow hemodynamic shifts) and residual temporal dependency between adjacent windows, buffer zones consisting of 50 segments (approximately 500 seconds, or 8–9 minutes) immediately before and after the test block were excluded from both training and testing. These gaps enforce strict temporal isolation between the training and test distributions, effectively suppressing the overly optimistic bias commonly observed in simple contiguous splitting schemes. Consequently, the proposed evaluation protocol establishes a challenging yet realistic setting that mimics prospective, real-world monitoring scenarios in which the model must generalize to unseen future hemodynamic states based solely on earlier observations. This design aligns with the most recent rigorous recommendations for avoiding temporal leakage in personalized blood pressure estimation studies~\cite{leitner2022, zhang2025personalized, leogrande2025dataset}, and the reported performance therefore provides a reliable estimate of resting-state personalized blood pressure estimation in everyday environments.
This procedure was repeated 10 times with different random seeds, and the results are reported as the mean and standard deviation. All models were trained using the Adam optimizer with a learning rate of $10^{-3}$, a batch size of 32, and a dropout rate of 0.5 for 100 epochs, using the mean squared error (MSE) loss. All experiments were implemented in PyTorch 2.0 and executed on Google Colab Pro equipped with an NVIDIA Tesla T4 GPU. Model performance was evaluated using two standard regression metrics, Mean Absolute Error (MAE) and Root Mean Squared Error (RMSE), computed separately for SBP and DBP:
\begin{equation}
\text{MAE} = \frac{1}{N} \sum_{i=1}^{N} |y_i - \hat{y}_i|, 
~\text{RMSE} = \sqrt{\frac{1}{N} \sum_{i=1}^{N} (y_i - \hat{y}_i)^2 }
\end{equation}
where $y_i$ and $\hat{y}_i$ denote the reference and estimated BP values for the $i^\text{th}$ sample. MAE measures the average prediction error, whereas RMSE penalizes large deviations, making it more sensitive to outliers.

As shown in Table~\ref{tab:hyper}, the fused CNN–morphology network was trained with Adam (lr = $1\times10^{-3}$) using an MSE loss for joint SBP/DBP regression, with batch size 32, 100 epochs, and dropout 0.5 applied to the regression head. Table~\ref{tab:ppg_baseline_features} summarizes a representative set of ten conventional morphology-based features that have been widely used in the PPG literature to characterize pulse shape and amplitude dynamics~\cite{Elgendi2012, Allen2007}. The first group of features captures the overall amplitude profile of the waveform, including the systolic peak level ($v_p$), the end-of-window amplitude ($v_t$), their difference ($d_v$), and the mean level over the segment ($v_m$). The second group describes dispersion and local dynamics: the standard deviation $\sigma_x$ reflects variability within the window, while the maximum upstroke slope $k_{\max}$ and its occurrence time $t_{k,\max}$ quantify the steepness and timing of the systolic rising edge. Finally, the area under the curve up to the peak $a_{\max}$, the normalized time to peak $t_{v_p}$, and the time at which the signal is closest to its mean level $t_m$ provide integral and temporal descriptors of the pulse, offering a compact summary of its energy distribution and temporal symmetry. In this work, these representative handcrafted features, together with their intuitive descriptions in Table~\ref{tab:ppg_baseline_features}, serve as a conventional morphological baseline against which the proposed feature set is quantitatively compared.

\begin{table}[!htbp]
\caption{Training configuration for the CNN branch model.}
\centering
\label{tab:hyper}
\begin{tabular}{@{}ll@{}}
\hline
Parameter & Value \\ \hline
Optimizer & Adam \\
Learning rate & $1\times10^{-3}$ \\
Loss & MSE (joint SBP/DBP regression) \\
Batch size & 32 \\
Epochs & 100 \\
Dropout & 0.5 \\
\hline
\end{tabular}
\label{tab:training_config}
\end{table}

\begin{table*}[!htbp]
\centering
/\footnotesize
\caption{Conventional morphological features extracted from a segment $x[n]$}
\label{tab:ppg_baseline_features}
\renewcommand{\arraystretch}{1.3}
\begin{tabular}{l | c | l | l}
\hline
Feature Name & Symbol & Mathematical Definition & Description \\
\hline
Peak amplitude 
& $v_p$ 
& $v_p = x[p]$ 
& Amplitude at the systolic peak \\
End amplitude 
& $v_t$ 
& $v_t = x[e]$ 
& Amplitude at the end of the window (baseline level) \\
Peak--end difference 
& $d_v$ 
& $d_v = v_p - v_t$ 
& Difference between peak and end amplitudes \\
Mean level 
& $v_m$ 
& $v_m = \frac{1}{N}\sum_{n=0}^{N-1} x[n]$ 
& Average signal level over the window \\
Standard deviation 
& $\sigma_x$ 
& $\sigma_x = \sqrt{\frac{1}{N}\sum_{n=0}^{N-1} (x[n] - v_m)^2}$ 
& Amplitude variability within the window \\
Max upstroke slope 
& $k_{\max}$ 
& $k_{\max} = \max_n (\Delta x[n])$ 
& Maximum rising slope during the systolic upstroke \\
Time of max slope 
& $t_{k,\max}$ 
& $t_{k,\max} = \frac{1}{N}\arg\max_n (\Delta x[n])$ 
& Normalized time of the maximum upstroke slope \\
Area up to peak 
& $a_{\max}$ 
& $a_{\max} \approx \sum_{n=0}^{p-1} x[n]$ 
& Approximate area under the curve from start to peak \\
Time to peak 
& $t_{v_p}$ 
& $t_{v_p} = \frac{p}{N}$ 
& Normalized time from start to peak \\
Time at mean level 
& $t_m$ 
& $t_m = \frac{1}{N}\arg\min_n |x[n] - v_m|$ 
& Time where the signal is closest to the mean level \\
\hline
\end{tabular}
\vspace{4pt}
\footnotesize
\\Note: $N$ is the window length, $p$ is the peak index, $e = N-1$, and $\Delta x[n]$ is the first difference.
\end{table*}

\FloatBarrier


\subsection{PERFORMANCE RESULTS}

\subsubsection{Effect of Prior Features in Personalized BP Estimation}

In the personalized setting, as shown in Table~\ref{tab:ablation_simple},  the CNN-only baseline achieved MAE values of 6.70 $\pm$ 1.02 mmHg for SBP and 3.49 $\pm$ 0.65 mmHg for DBP. Augmenting the CNN with conventional morphological features already led to a substantial improvement, reducing the errors to 3.99 $\pm$ 0.48 mmHg (SBP) and 2.78 $\pm$ 0.61 mmHg (DBP). When these conventional features were replaced by the proposed morphological priors, the MAE further decreased to $3.77 \pm 0.50$ mmHg for SBP and $2.36 \pm 0.40$ mmHg for DBP, corresponding to relative reductions of approximately 44\% and 32\% compared with the CNN-only baseline, respectively. These results confirm that explicitly encoding person-specific morphological priors provides complementary information to the raw PPG representation and significantly enhances personalized cuffless BP estimation.

\begin{table*}[!htbp]
\centering
\footnotesize
\caption{Ablation study on the effect of incorporating conventional and proposed features into the CNN model for personalized BP estimation. The MAE is reported as mean $\pm$ standard deviation over 5 folds, and the improvement over the personalized CNN-only baseline is statistically significant ($p < 0.05$, paired t-test).}
\label{tab:ablation_simple}
\begin{tabular}{lccll}
\hline
\textbf{Configuration} & \textbf{MAE$_\mathrm{SBP}$} & \textbf{MAE$_\mathrm{DBP}$} & \textbf{Params (K)} & \textbf{FLOPs (M)}
\\
\hline
CNN Only (Baseline) & 6.70 $\pm$ 1.02 & 3.49 $\pm$ 0.65 & 60.0 & 3.12 \\
CNN + Conventional Features  & 3.99 $\pm$ 0.48 & 2.78 $\pm$ 0.61 & 101.9 & 3.16\\
\textbf{CNN + Proposed Features (Ours)}& \textbf{3.77} $\pm$ \textbf{0.50} & \textbf{2.36} $\pm$ \textbf{0.40} & \textbf{101.9 }& \textbf{3.16}\\ \hline
\end{tabular}
\end{table*}

\FloatBarrier

At the same time, the proposed fusion model remains highly lightweight. Incorporating the prior features increases the model size only modestly from 60.0 K to 101.9 K parameters, while the computational cost grows marginally from 3.12 M to 3.16 M FLOPs per inference. This small footprint and low computational burden indicate that the proposed architecture achieves improved accuracy without sacrificing efficiency, making it suitable for real-time deployment on resource-constrained wearable and mobile devices. While recent advancements in deep learning, such as Transformer-based architectures, may offer marginal improvements in predictive accuracy, they typically incur a massive computational overhead that prohibits continuous, on-chip inference. In contrast, our hybrid model requires only 3.16 M FLOPs—orders of magnitude lower than attention-based models—demonstrating that strategic feature engineering can achieve high-performance BP estimation within the strict power and thermal constraints of commercial wearable microcontrollers.

\subsubsection{Effect of Prior Features in Cross-Subject BP Estimation}

Although the proposed framework is primarily designed for personalized BP estimation, we additionally evaluated the same CNN backbone and feature configurations in a more challenging cross-subject setting to verify the generality of the prior features. Specifically, data from 100 subjects were partitioned into 5 folds in a subject-independent manner, where in each fold a disjoint subset of subjects was held out for testing while the remaining subjects were used for training. As summarized in Table~\ref{tab:ablation_cross}, incorporating conventional handcrafted features into the CNN slightly reduces the cross-subject MAE compared with the CNN-only baseline. Furthermore, augmenting the CNN with the proposed prior features yields a modest yet statistically significant reduction in both SBP and DBP errors ($p < 0.05$), from 9.14 to 8.73~mmHg for SBP and from 5.01 to 4.52~mmHg for DBP. It should be emphasized that these error levels are still insufficient for practical or clinical-grade cuffless BP estimation; the cross-subject experiment is therefore reported solely as a stress test to assess the subject-invariant contribution of the proposed prior features, rather than as a deployable cross-subject BP estimation setting.

\begin{table}[!htbp]
\centering
\footnotesize
\caption{Ablation study on the effect of incorporating conventional and proposed features into the CNN model for cross-subject BP estimation. The MAE is reported as mean $\pm$ standard deviation over 5 folds, and the improvement over the cross-subject CNN-only baseline is statistically significant ($p < 0.05$, paired t-test).}
\label{tab:ablation_cross}
\begin{tabular}{lcc}
\hline
\textbf{Configuration} & \textbf{MAE$_\mathrm{SBP}$} & \textbf{MAE$_\mathrm{DBP}$} \\
\hline
CNN Only (Baseline)
& 9.14 $\pm$ 1.14 
& 5.01 $\pm$ 1.03 \\
CNN + Conventional Features 
& 9.03 $\pm$ 1.02 
& 4.97 $\pm$ 1.10 \\
\textbf{CNN + Proposed Features (Ours)}
& \textbf{8.73 $\pm$ 1.15 }
& \textbf{4.52 $\pm$ 1.20} \\
\hline
\end{tabular}
\end{table}

\FloatBarrier

\subsubsection{Bias and Agreement Assessment}

\begin{figure}[!htbp]
\centering
\includegraphics[width=0.6\linewidth]{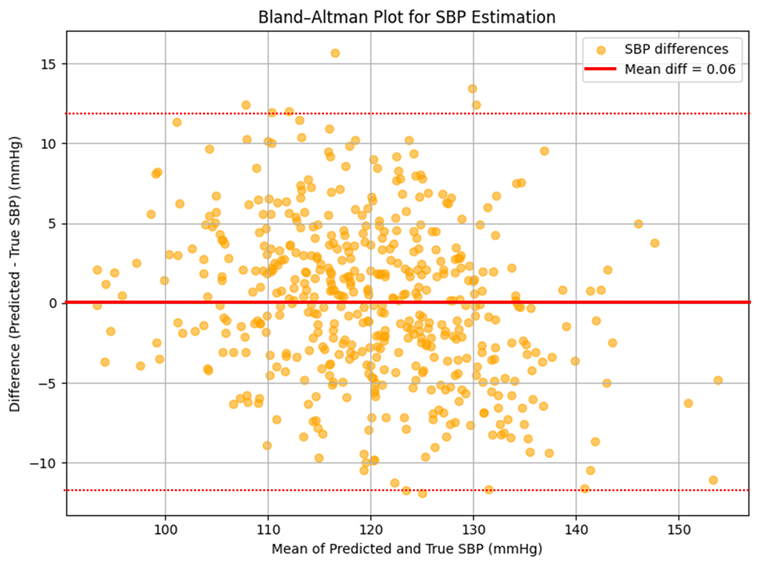}
\caption{
Bland--Altman analysis for SBP estimation showing a mean difference of $+0.06\,\mathrm{mmHg}$ (red solid line) and the 95\% limits of agreement (LoA) of $-12$ to $+12\,\mathrm{mmHg}$ (red dotted lines), indicating acceptable agreement with minimal systematic bias. Note that the plot is generated on a per-segment basis; points at higher SBP levels may correspond to transiently elevated segments from otherwise normotensive participants.
}
\label{fig:ba_sbp}
\end{figure}

\begin{figure}[!htbp]
\centering
\includegraphics[width=0.6\linewidth]{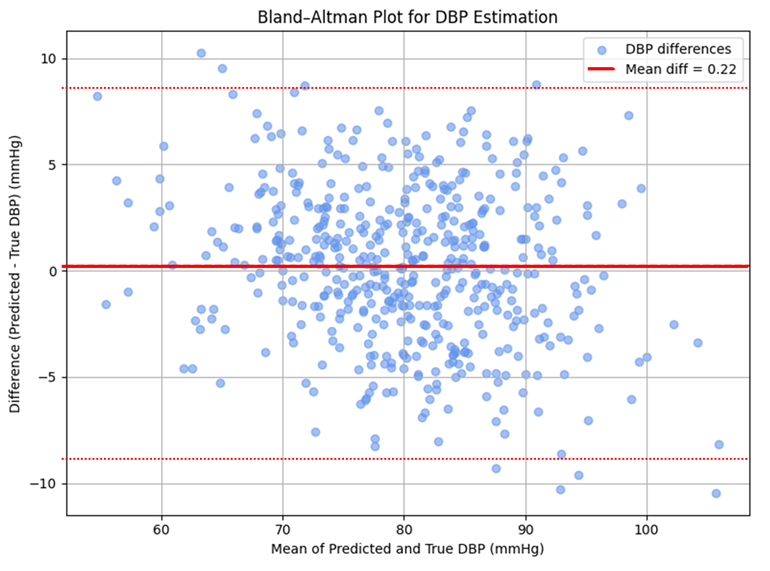}
\caption{
Bland--Altman analysis for DBP estimation showing a mean difference of $+0.22\,\mathrm{mmHg}$ (red solid line) and the 95\% limits of agreement (LoA) of $-9$ to $+9\,\mathrm{mmHg}$ (red dotted lines), indicating acceptable agreement with minimal systematic bias. Because the analysis is also segment-based, points at higher DBP levels may reflect transiently elevated segments within normotensive subjects.
}
\label{fig:ba_dbp}
\end{figure}

The Bland–Altman plots in Fig. ~\ref{fig:ba_sbp} and Fig.~\ref{fig:ba_dbp} show negligible bias for both SBP and DBP estimates. While a few outliers are observed, the majority of points fall within clinically acceptable limits, supporting the accuracy of the proposed model under the tested conditions.

\FloatBarrier

\subsubsection{Comparison with Recent Personalized BP Estimation Methods}

As shown in Table~\ref{tab:comparison_year}, our proposed approach achieves personalized BP estimation accuracy comparable to or better than state-of-the-art methods that rely on large-scale pretraining, subject-specific fine-tuning, or incremental personalization. Unlike these transfer learning or meta-learning-based strategies—which often require extensive subject data or complex adaptation steps—our model obtains robust individualization simply by incorporating morphological priors, without any additional pretraining or post-hoc calibration. Although variations in datasets and evaluation protocols prevent strict one-to-one comparison, these results strongly support the practicality of our approach for scalable, user-friendly personalized BP monitoring.

\begin{table*}[!htbp]
\centering
\scriptsize
\caption{Comparison with representative personalized BP estimation methods.}
\label{tab:comparison_year}
\begin{tabular}{l l l c c}
\hline
\textbf{Year} & \textbf{Study} & \textbf{Dataset} & \textbf{Personalization Approach} & \textbf{MAE (SBP / DBP)} \\
\hline
2022 & Leitner et al.\ Baseline~\cite{leitner2022} & MIMIC-III Subset & None & 4.59 / 2.72 \\
2022 & Leitner et al.\ Transfer~\cite{leitner2022} & MIMIC-III Subset & Pretraining + Fine-tuning & 3.52 / 2.20 \\
2023 & Fan et al.~\cite{fan2023fewbp} & MIMIC-II + Real-world & Pretraining + Adapter & 6.68 / 3.91 \\
2022 & Zhao et al.~\cite{zhao2022lightweight} & MIMIC-III Subset & Pretraining + Fine-tuning & 8.44 / 8.26 \\
2024 & Wang et al.~\cite{wang2024efisvr} & Real-world (IoT) & Personalized incremental learning & 3.11 / 2.47 \\
2024 & Guo et al.~\cite{guo2024meta} & Multi-domain & Meta-learning + Fine-tuning & 7.46 / 4.51 \\
2025 & Zhang et al.~\cite{zhang2025personalized} & PPG-BP & Pretraining + Calibration & 7.08 / 5.53 \\
2025 & \textbf{This work} & MIMIC-III Subset & None & \textbf{3.77} $\pm$ \textbf{0.50} / \textbf{2.36} $\pm$ \textbf{0.40} \\
\hline
\end{tabular}
\end{table*}

\FloatBarrier


\subsection{Explainability Analysis via SHAP}

\begin{figure}[!htbp]
  \centering
  \includegraphics[width=0.6\linewidth]{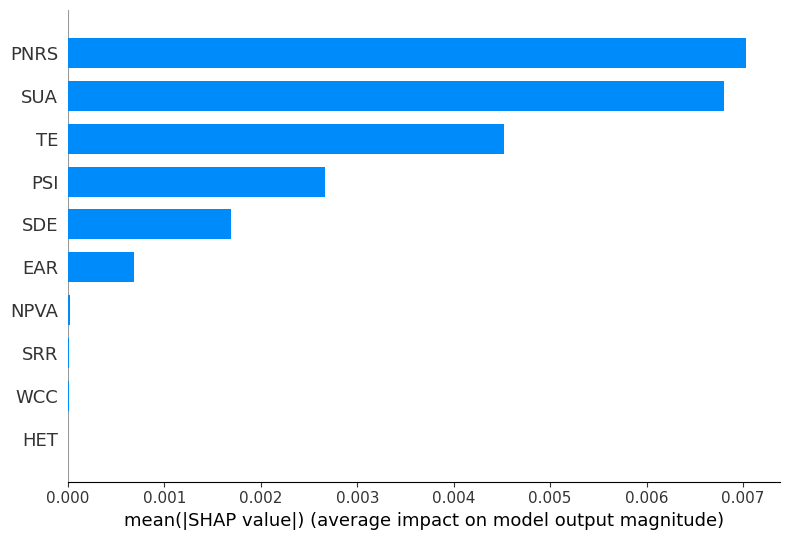}
  \caption{Global SHAP summary plot. PNRS and SUA exhibit the largest contributions, followed by TE, while PSI and SDE have moderate effects. WCC, EAR, SRR, NPVA, and HET exhibited relatively low SHAP values, suggesting that their marginal contribution may stem from redundancy with dominant features (e.g., PNRS and SUA) or reduced variance across the study cohort. Despite their lower rankings, these features may still capture secondary physiological cues in broader populations or under stress conditions.}
  \label{fig:shap_summary}
\end{figure}

To interpret the contribution of handcrafted PPG features to SBP prediction, SHAP were applied. The global SHAP summary plot (Fig.~\ref{fig:shap_summary}) ranks PNRS, SUA, and TE as the most influential features for SBP prediction, followed by PSI and SDE with moderate contributions. The remaining features, including WCC, EAR, SRR, NPVA, and HET, exhibit minimal impact. This ranking indicates that a small subset of morphological features accounts for the majority of the model’s predictive power.

\FloatBarrier

\begin{figure}[!htbp]
  \centering
  \includegraphics[width=0.6\linewidth]{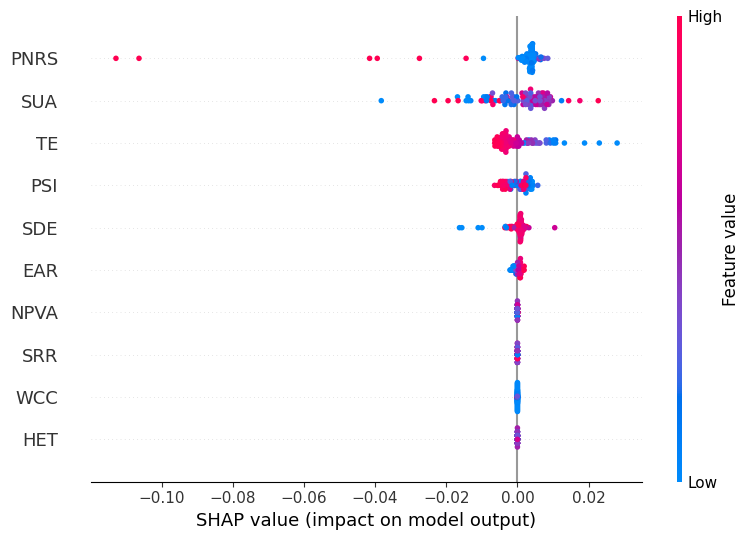}
  \caption{Mean absolute SHAP value ranking. PNRS and SUA dominate the model's predictions, TE is moderately influential, and PSI and SDE provide secondary contributions.}
  \label{fig:shap_bar}
\end{figure}

\begin{figure}[!htbp]
  \centering
  \includegraphics[width=0.6\linewidth]{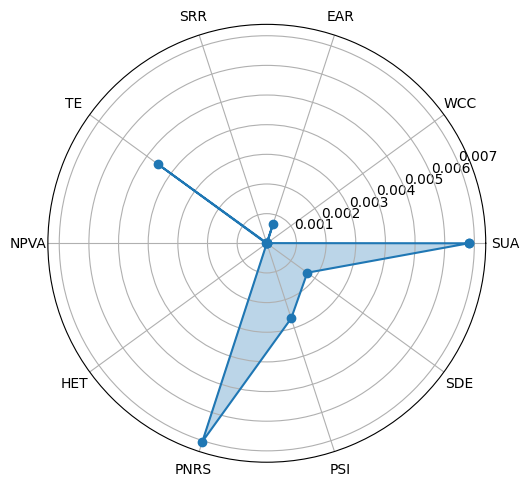}
  \caption{Radar chart of normalized SHAP values. PNRS and SUA form the largest wedges, with TE, PSI, and SDE contributing moderately. Remaining features are near-zero.}
  \label{fig:shap_radar}
\end{figure}

The mean absolute SHAP values (Fig.~\ref{fig:shap_bar}) corroborate these findings, ranking PNRS and SUA as dominant, TE as moderately important, and PSI and SDE as secondary contributors. A normalized radar visualization (Fig.~\ref{fig:shap_radar}) further confirms the relative influence hierarchy, with PNRS and SUA forming the largest feature wedges.

\FloatBarrier

\begin{figure}[!htbp]
  \centering
  \includegraphics[width=0.6\linewidth]{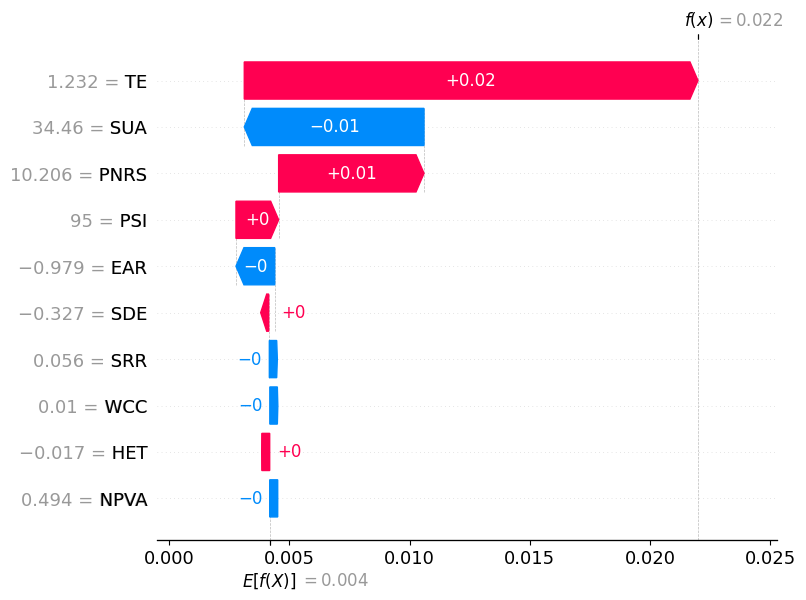}
  \caption{SHAP waterfall for segment \#1. High PNRS and TE drive positive shifts, while SUA slightly decreases prediction.}
  \label{fig:shap_wf1}
\end{figure}

\begin{figure}[!htbp]
  \centering
  \includegraphics[width=0.6\linewidth]{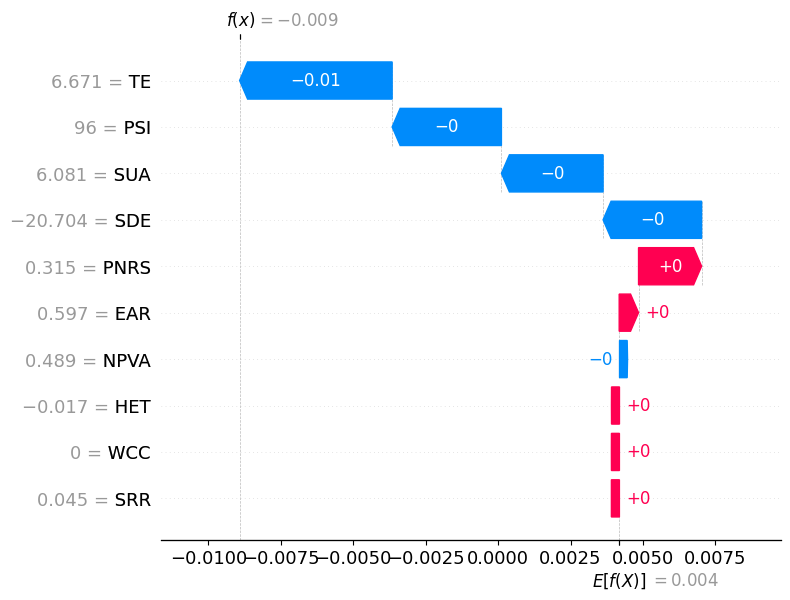}
  \caption{SHAP waterfall for segment \#2. Negative TE dominates downward influence; PNRS provides a positive offset.}
  \label{fig:shap_wf2}
\end{figure}

\begin{figure}[!htbp]
  \centering
  \includegraphics[width=0.6\linewidth]{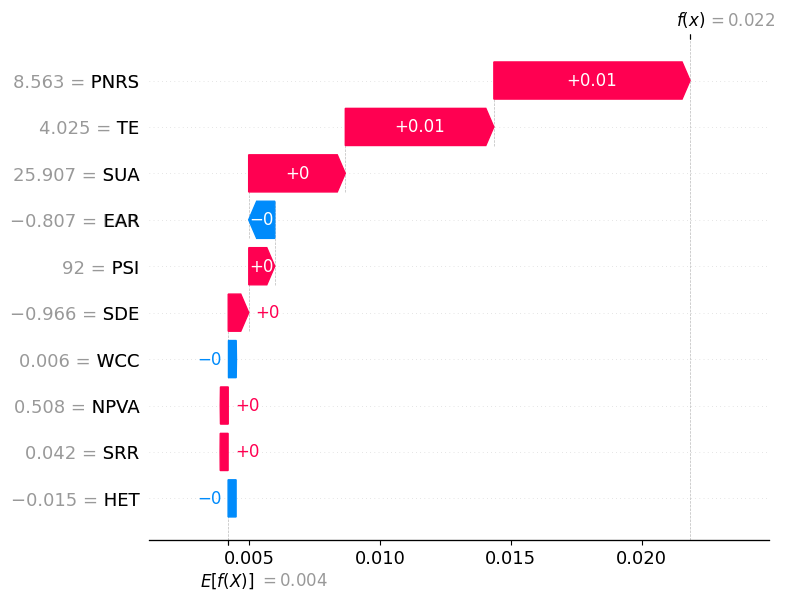}
  \caption{SHAP waterfall for segment \#3. PNRS and TE again lead positive contributions, confirming global ranking trends.}
  \label{fig:shap_wf3}
\end{figure}

Instance-level SHAP waterfall plots (Figs.~\ref{fig:shap_wf1}–\ref{fig:shap_wf3}) provide local interpretability. For segment \#1, PNRS and TE strongly drive positive prediction shifts, with SUA exerting a mild negative offset. In segment \#2, negative TE is the primary downward driver, partially counterbalanced by PNRS. Segment \#3 again highlights PNRS and TE as the dominant positive influencers, confirming their consistency across samples.

\FloatBarrier

Collectively, these analyses demonstrate that PNRS, SUA, and TE are the most influential predictors of SBP, respectively reflecting reflective waves, systolic energy, and waveform irregularity. Their consistent dominance in both global and local analyses enhances the model’s interpretability and underscores potential clinical relevance, warranting further validation with clinical experts. 


\subsection{Physiological Interpretation of SHAP Patterns}

SHAP analysis highlighted morphological features with clear physiological relevance. PNRS reflects arterial stiffness and the magnitude of reflective waves, both contributing to systolic pressure augmentation. SUA represents cumulative systolic upstroke energy, linking elevated values to reduced arterial compliance. TE quantifies diastolic-phase waveform irregularity, potentially indicating impaired vascular regulation. PSI and SDE capture secondary characteristics such as systolic slope and curvature, which relate to myocardial contractility and vascular tone. In contrast, features such as WCC, EAR, SRR, NPVA, and HET showed lower contributions in this dataset, likely due to limited variability and partial redundancy with more dominant features such as PNRS and SUA. By embedding these physiologically interpretable descriptors into the model, it is possible to connect data-driven predictions with established cardiovascular mechanisms, enhancing transparency and clinical interpretability. These interpretations are specific to the present dataset and model configuration and may not generalize to other populations or measurement conditions. Moreover, SHAP analysis quantifies statistical contribution rather than causal influence, underscoring the need for external validation across diverse cohorts and long-term monitoring scenarios.

\subsection{Discussion} 

Although the proposed hybrid CNN-morphological feature-based personalized blood pressure estimation framework demonstrated superior performance and physiological interpretability, several limitations warrant discussion. A primary limitation is the reliance on a single clinical dataset (MIMIC-III). While the focus on normotensive subjects aligns with the preventative monitoring scope defined in Section IV, this restricts the model’s immediate applicability to clinical hypertension management, where vascular stiffening and waveform alteration are more pronounced.

Furthermore, regarding validation on external datasets, this study prioritized a rigorous cross-subject validation protocol within the PulseDB/MIMIC-III cohort rather than incorporating heterogeneous external datasets. This strategic decision was made to isolate physiological variability from sensor-specific discrepancies—specifically the domain shift between multi-wavelength and standard PPG sensors—thereby ensuring that the reported performance reflects the model's capability to learn true hemodynamic patterns rather than sensor artifacts. Nevertheless, we acknowledge that clinical generalizability requires validation beyond a controlled single-source database. Although we employed a strict chronological block split to mitigate data leakage risks, the model's robustness against complex motion artifacts and environmental noise inherent to daily ambulatory settings remains to be fully verified. Consequently, future studies must extend validation to external datasets with compatible sensor modalities and conduct longitudinal trials in real-world environments to fully establish the framework's reliability for ubiquitous health monitoring.

\section{Conclusion}
This work introduced a personalized BP estimation framework that requires no pretraining or post-hoc calibration and is evaluated in a subject-specific protocol, achieving 3.77 mmHg (SBP) and 2.36 mmHg (DBP) while providing interpretable feature contributions. By embedding a handcrafted morphological feature set into a hybrid CNN architecture, the proposed model captures both localized and global PPG waveform dynamics while maintaining robustness to signal variability. Unlike conventional feature-based methods, which depend heavily on fiducial point accuracy, our approach minimizes such dependency and incorporates clinically meaningful descriptors related to arterial compliance and cardiac contractility. Evaluation on the MIMIC-III dataset demonstrated that the proposed model achieves MAEs of 3.77 mmHg for SBP and 2.36 mmHg for DBP, which are comparable to previously reported methods under similar evaluation settings.

Bland--Altman analysis confirmed strong agreement with reference measurements, and SHAP-based interpretation provided valuable insights into feature contributions, bridging the gap between predictive performance and clinical transparency. Future work will explore adaptive calibration strategies and multi-domain feature integration to further improve generalization under real-world conditions. These findings highlight the potential of interpretable, domain-informed neural architectures for advancing research on continuous and personalized BP monitoring. Further validation in diverse, real-world wearable environments is warranted.

\section*{Acknowledgments}
This work was supported by the Institute of Information \& Communications Technology Planning \& Evaluation(IITP) grant funded by the Korea government(MSIT) (No.RS-2024-00357879, AI-based Biosignal Fusion and Generation Technology for Intelligent Personalized Chronic Disease Management)

\end{document}